\documentclass[sigchi]{acmart} 

\usepackage[inline]{enumitem}
\usepackage{hyperref}
\def\BibTeX{{\rm B\kern-.05em{\sc i\kern-.025em b}\kern-.08emT\kern-.1667em\lower.7ex\hbox{E}\kern-.125emX}}
    
\setcopyright{rightsretained} 
\copyrightyear{2019}
\acmYear{2019}
\setcopyright{acmlicensed}
\acmConference[EICS '19]{EICS '19: ACM SIGCHI Symposium on Engineering Interactive Computing Systems}{June 18-21, 2019}{Valencia, Spain}
\acmBooktitle{Valencia '19: ACM SIGCHI Symposium on Engineering Interactive Computing Systems, June 18-21, 2019, 2018, Valencia, Spain}
\acmPrice{15.00}
\acmDOI{10.1145/1122445.1122456}
\acmISBN{978-1-4503-9999-9/18/06}

\begin{document}

%
\title{A System for Real-Time Interactive Analysis of Deep Learning Training}

%
\author{Shital Shah}
\affiliation{%
  \institution{Microsoft Research}
  \streetaddress{1 Microsoft Way}
  \city{Redmond}
  \state{Washington}
  \postcode{98052}
}
\email{shitals@microsoft.com}

\author{Roland Fernandez}
\affiliation{%
  \institution{Microsoft Research}
  \streetaddress{1 Microsoft Way}
  \city{Redmond}
  \state{Washington}
  \postcode{98052}
}
\email{rfernand@microsoft.com}

\author{Steven Drucker}
\affiliation{%
  \institution{Microsoft Research}
  \streetaddress{1 Microsoft Way}
  \city{Redmond}
  \state{Washington}
  \postcode{98052}
}
\email{sdrucker@microsoft.com}

%

%
\begin{abstract}
  Performing diagnosis or exploratory analysis during the training of deep learning models is challenging but often necessary for making a sequence of decisions guided by the incremental observations. Currently available systems for this purpose are limited to monitoring only the logged data that must be specified before the training process starts. Each time a new information is desired, a cycle of stop-change-restart is required in the training process. These limitations make interactive exploration and diagnosis tasks difficult, imposing long tedious iterations during the model development. We present a new system that enables users to perform interactive queries on live processes generating real-time information that can be rendered in multiple formats on multiple surfaces in the form of several desired visualizations simultaneously. To achieve this, we model various exploratory inspection and diagnostic tasks for deep learning training processes as specifications for streams using a map-reduce paradigm with which many data scientists are already familiar. Our design achieves generality and extensibility by defining composable primitives which is a fundamentally different approach than is used by currently available systems. The open source implementation of our system is available as TensorWatch project at https://github.com/microsoft/tensorwatch.
\end{abstract}

%
%


%

%

\copyrightyear{2019} 
\acmYear{2019} 
\acmConference[EICS '19]{ACM SIGCHI Symposium on Engineering Interactive Computing Systems}{June 18--21, 2019}{Valencia, Spain}
\acmBooktitle{ACM SIGCHI Symposium on Engineering Interactive Computing Systems (EICS '19), June 18--21, 2019, Valencia, Spain}\acmDOI{10.1145/3319499.3328231}
\acmISBN{978-1-4503-6745-5/19/06}

%
\setlist[itemize]{noitemsep, topsep=0pt}
\setlist[enumerate]{noitemsep, topsep=0pt}

\maketitle

\section{Introduction}
The rise of deep learning is complimented by ever increasing model complexity, size of the datasets, and corresponding longer training times to develop the model. For example, finishing 90-epoch ImageNet-1k training with ResNet-50 takes 14 days to complete on an NVIDIA M40 GPU\cite{You:2018:ITM:3225058.3225069}. Researchers and practitioners often find themselves losing productivity due to the inability to quickly obtain desired information dynamically from the training process without having to incur stop-change-restart cycles.  While a few solutions have been developed for real-time monitoring of deep learning training, there has been a distinct lack of systems that offer dynamic expressiveness through conversational style interactivity supporting the exploratory paradigm.

In this paper we offer a new system that enables the dynamic specification of queries and eliminates the requirement to halt the learning process each time a new output is desired. We also enable the displays of multiple, simultaneous visualizations that can be generated on-demand by routing to them the desired information chosen by the user. The pillars of our architecture are general enough to apply our system design to other domains with similar long running processes.

Our main contributions are
\begin{enumerate*}
\item system design based on dynamic stream generation using map-reduce as the Domain Specific Language (DSL) to perform interactive analysis of long running processes such as machine learning training
\item separation of concerns that allows to build dynamic stream processing pipeline with visualizations agnostic of rendering surfaces, and 
\item abstraction to allow the comparison of previously generated heterogeneous data along with the live data in a desired set of visualizations, all specified at the runtime.
\end{enumerate*}

\section{Related Work}
TensorBoard\cite{Wongsuphasawat2018} is currently among the most popular of monitoring tools, offering a variety of capabilities including data exploration using dimensionality reduction, and model data flow graphs. However, its monitoring capabilities are limited to viewing only the data that was explicitly specified to be logged before the training starts. While the tool provides some interactivity in visualization widgets, no  interactivity is provided in terms of dynamic queries. Furthermore, few pre-defined visualizations are offered in the dashboard in a pre-configured tabbed interface and thus somewhat limited in other layout preferences. 

The logging-based model is also used by other frameworks, including Visdom\cite{Choo2018} and VisualDL\cite{VisualDL}. Several authors\cite{Liu2017,DBLP:journals/corr/abs-1712-05902,Choo2018} have identified the research opportunities in diagnostic aspects of deep learning training and interactively analyzing it due to time consuming trial-and-error procedures.

Map-reduce is an extensively studied paradigm originated from the functional programming \cite{Steele1995} and successfully utilized for constructing data flows and performing large scale data processing in the field of distributed computing \cite{Dean2008,Gates:2009:BHD:1687553.1687568,Catanzaro2008AMR}. Many variants of map-reduce has been created\cite{Afrati:2011:MER:1951365.1951367} for a variety of scenarios, and it has also gained wide adoption for the various data analysis tasks\cite{Ekanayake2008,Pavlo:2009:CAL:1559845.1559865}.

Visualizations based on streams have been studied deeply for the real-time data scenarios\cite{(Ed.)98datavisualization,Traub2017} with various systems aiming to enhance interactivity, adaptability, performance and dynamic configurations\cite{Logre:2018:MSV:3233739.3229096,ellis2014real, Roberts2007, Few:2006:IDD:1206491}. Query driven visualizations have been popular for databases utilizing SQL and big data using custom DSLs\cite{Stockinger, Babu:2001:CQO:603867.603884, Plale2003}. This paradigm also becomes a cornerstone in our system design.


\section{Scenarios}
We describe a few real-world scenarios in this section to develop an intuition of the requirements and understanding of problems often faced by the practitioners.

\subsection{Diagnosing Deep Learning Training}
John is the deep learning practitioner with the task of developing a model for gender identification from a large labeled dataset of human faces. As each experiment takes several minutes even on a reduced subset of the data, John wishes to view training loss and accuracy trends in real time. In many experiments, training loss does not seem to be reducing and to understand the cause John needs to view the gradient flow chart. However, this requires John to terminate the training process, add additional logging for this information, and then restart the training. As John observes the gradient flow chart, he starts suspecting that his network may be suffering from the vanishing gradient problem. To be sure, John now wishes to view growth of weights and the distribution of initial values. This new information again causes a stop-change-restart cycle adding significant cost to obtain each new piece of information in the diagnostic process that is inherently iterative.

\subsection{Analyzing The Model Interpretation Results}
Susan is using a GAMs framework\cite{Hastie1986} to analyze the impact of each feature in her model. As a data scientist, she depends on Jupyter Notebook to perform her analysis. As the computation takes several minutes before generating the desired charts, Susan wants to display progressive visualizations displaying partial results as they evolve. Instead of designing and implementing a custom system for her one-off experiment, it would be ideal if she could  simply generate a stream of data using map-reduce paradigm that she is already familiar with. This stream can then be easily painted to the desired rendering surface such as Jupyter Notebook to display progressive real-time visualization as her models evolve in time. 

\subsection{Diagnosing and Managing Deep Learning Jobs}
Rachel spins up several dozens of deep learning jobs as part of her experiments in GPU cloud infrastructure. These long running jobs may take many days to complete and therefore are expensive to run in cloud. However, it turns out that many of the poorer performing jobs could be identified much earlier and be terminated, thus freeing up the expensive resources. However, designing and building such infrastructure is time consuming and requires additional engineering skills. It would be ideal for Rachel if she could simply output streams of performance data from her jobs and then create a small monitoring application that consumes these streams.

\section{System Design}

\begin{figure*}[ht]
  \centering
  \includegraphics[height=3.5in]{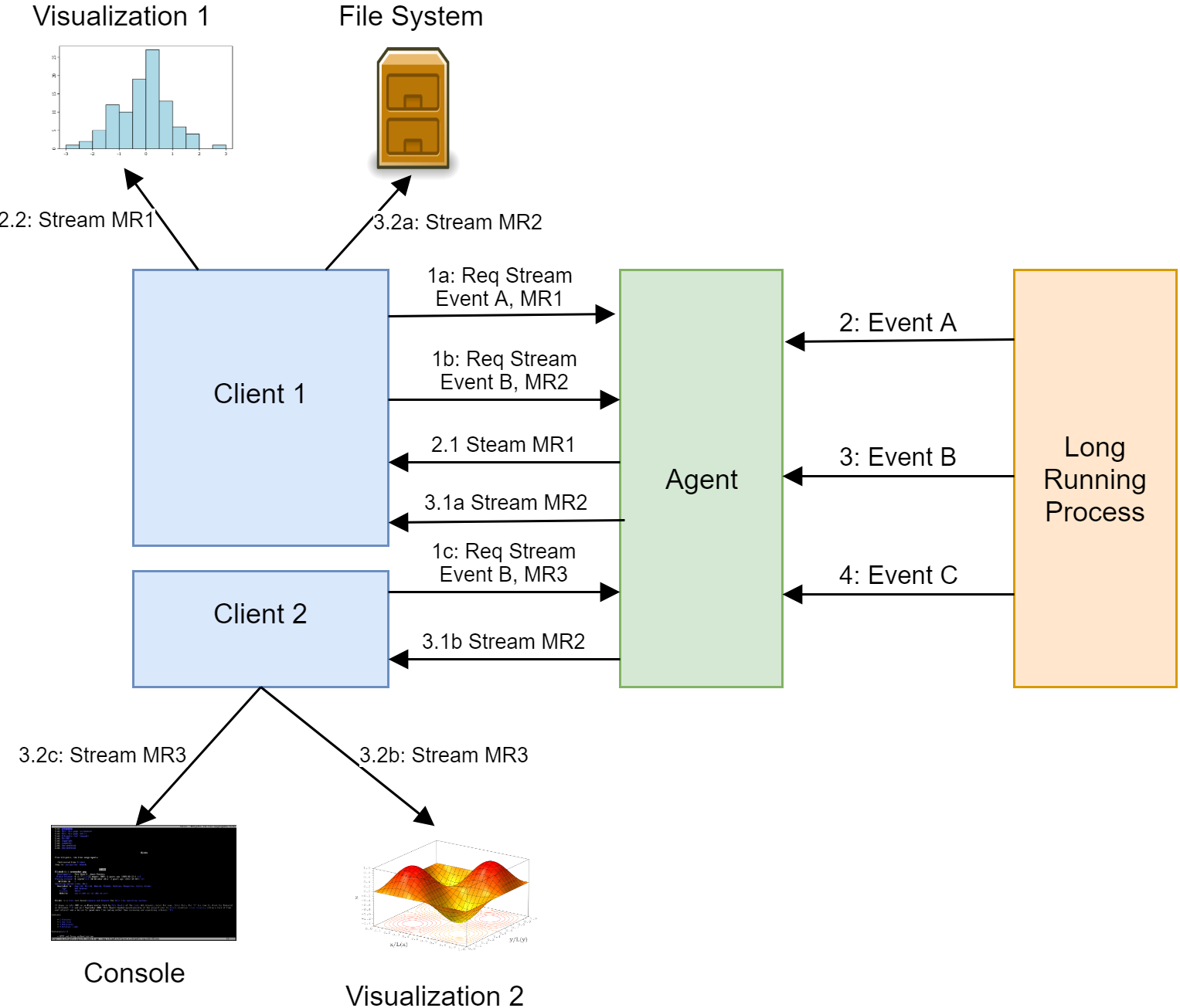}
  \caption{Collaboration diagram for our system depicting interactions between various actors. Standard notations are used with numbered interactions indicating their sequence with alphabet suffix denoting the potential concurrency. Our system includes the long running process generating various events, clients making requests for stream using map-reduce queries (denoted by MRx) for the desired events and the agent responding back with resultant streams that can be directed to desired visualizations or other processes.}
  \label{fig:TensorWatch_Collaboration}
\end{figure*}

\subsection{Key Actors}
Our system design contains the following key actors as shown in Figure \ref{fig:TensorWatch_Collaboration}:
\begin{enumerate}
  \item A long running process $P$ such as a deep learning training process.
  \item Zero or more clients that may be located on the  same or different machines as $P$.
  \item An agent $A$ that is embedded in $P$ listening to requests from the clients
\end{enumerate}
\subsection{The Long Running Process}
We abstract three specific characteristics of a long running process $P$:
\begin{enumerate}
  \item $P$ may generate many types of events during its lifetime. Each type of event may occur multiple times, but the sequence of events is always serialized, i.e., there is never more than one event of the same type occurring at the same time in the same process.
  \item As events of each type are strictly ordered so that we can optionally assign a group to any arbitrary contiguous set of events. This ability will enable windowing for the reduce operator discussed later.
  \item For each event, optionally a set of values may be available for the observation. For example, on a batch completion event the metrics for that batch may be available for the observation. The process informs the agent when an event occurs and provides access to these observables.
\end{enumerate}
\subsection{The Client}
At any point in time multiple clients may exist simultaneously issuing the queries and consuming corresponding resultant streams. Each query can be viewed as a stream specification with the following attributes:
\begin{enumerate}
  \item The event type for which a stream should be generated. An event type may have multiple associated streams but each stream has only one associated event type.
  \item An expression in the form of map-reduce operations. This expression is applied to the observables at the time of event and the output becomes the value in the resultant stream.
\end{enumerate}

The client may utilize the resultant stream by directing it to multiple processes such as visualizations chosen at the runtime. Thus the same stream may generate a visualization as well as become input to another process in the data flow pipeline.

\subsection{The Agent}
The agent runs in-process in the host long running process $P$ and characterized by the following responsibilities:

\begin{enumerate}
  \item Listening to incoming requests for the creation of a stream. This is done asynchronously without blocking the host process $P$.
  \item When $P$ informs the agent that an event has occurred, the agent determines if any active streams exist for that event. If so, the agent executes the map-reduce computation attached to each stream for that event and sends the result of this computation back to the client.
\end{enumerate}

An important aspect of the agent design is that if there are no streams requested for an event then there is almost no performance penalty. Also, there is no performance penalty for having access to the large numbers of observables. This means that user may specify all of the variables of interest as observables beforehand and later use queries to use subset of them depending on the task.

\subsection{Example: Implementation for Deep Learning Training}

As an example of how above actors and abstractions may be utilized, consider the deep learning training scenario. This process performs computation in series of epochs, completion of each is an \emph{epoch event}. During each epoch, we execute several batches of data, completion of each becoming a \emph{batch event}. At each batch event we may observe the metric object that contains several statistics for the batch. Contiguous set of batch events within each epoch can be treated as one group. At the end of an epoch, we may want to compute some aggregated statistics which can easily be done by specifying the map-reduce expression that extracts the desired value from the metric object and performing the aggregation operation on it.

\subsection{Multiple Processes and Streams}
The above abstractions can easily be utilized to efficiently inspect many simultaneously running processes and make decisions such as early termination or modify desired parameters at the runtime. A user can also compare and visualize arbitrarily chosen subsets of jobs. 

\subsection{Modifying the State of a Long Running Process}
Our design trivially enables a useful capability of changing the observables of the long running process. In the context of deep learning training, this can be used for interactive hyper parameter tuning guided by observations \cite{NIPS2011_4443}. We simply allow users to send commands from interfaces such as Jupyter Notebook to the agent running in the host process. The agent then executes these commands on observables presented to it by the host process at the specified events.

\subsection{Stream Persistence}
One of the significant disadvantages of many current systems is the requirement that data of interest must be logged to the disk storage, which can become an expensive bottleneck. Our design with stream abstraction trivially enables pay-what-you-use model so that users can selectively specify at runtime to persist only those streams that they may be interested in viewing or comparison in the future.

\section{Stream Visualization}
Once a stream is produced, it can be visualized, stored or processed further in users data flow graph. 

\begin{figure}[h]
  \centering
  \includegraphics[width=\linewidth]{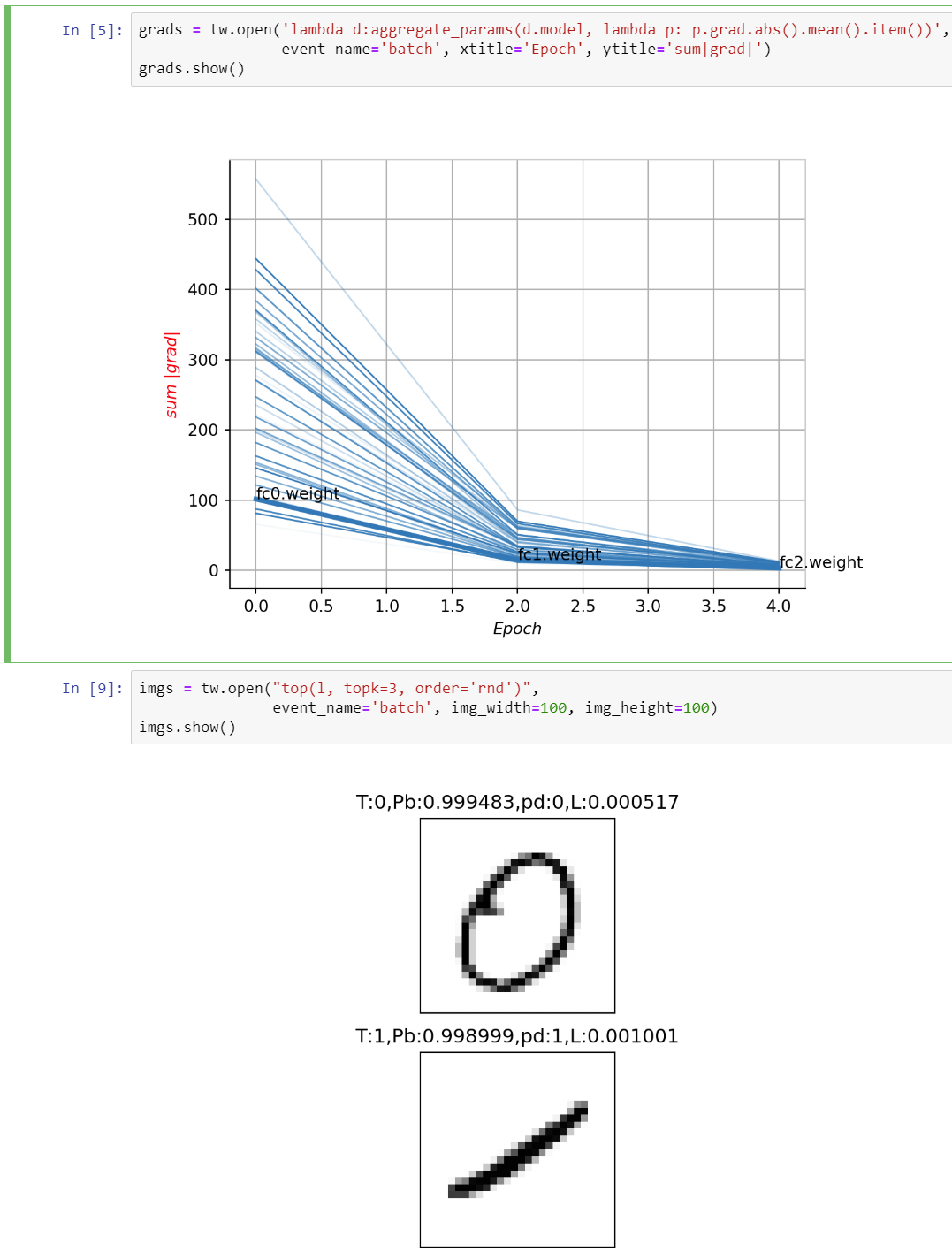}
  \includegraphics[width=\linewidth]{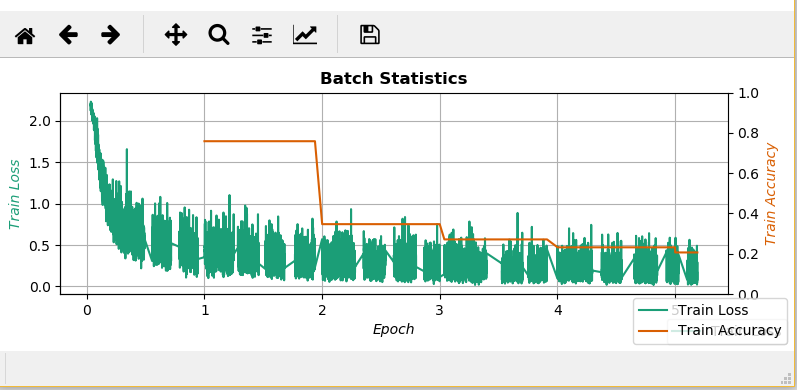}  
  \caption{Screenshot of three simultaneous real-time visualizations on two different surfaces generated dynamically by an user for the MNIST training process. On the top is an interactive session in Jupyter Notebook where the user specifies map-reduce queries in a cell for each desired visualization. The first output cell at the top shows evolution of average absolute gradients for each layer with lighter lines indicating the older plots. The second output cell shows random sample of predictions so far. At the bottom is the plot of two batch statistics rendered in a separate native application.}  
\end{figure}

\subsection{Adaptive Visualizers}
As we allow users to generate arbitrary streams dynamically, it becomes important that visualization widgets are specifically designed for automatic configuration by reflecting on data available in the stream. We adopt the adaptive visualization paradigm\cite{Nazemi2016, mourlas2009intelligent} for this purpose. For example, a visualizer may decide to paint a stream that has a tuple of two numeric values as a 2D line chart, tuple of 3 numeric values as a 3D line chart and tuple of 2 numeric and 1 string value as annotated 2D line chart. The user may override to select a precise rendering for a given stream.

A visualizer may allow adding or removing streams dynamically. The streams may not have the same data type allowing for the heterogeneous visualizations such as display of a histogram and a line chart overlays. If a visualizer receives incompatible streams than it may display an error. In the context of deep learning, this enables capabilities such as viewing multiple related metrics in the same visualization or comparing data generated by multiple experiments in the same visualization.



\subsection{Frame Based Animated Visualizations}
Many useful visualizations may consume values in a stream one after another as they arrive, e.g.\ , line charts. Another interesting scenario is to consider each value in the stream providing the complete data for each \emph{frame} in the visualization. This enables users to create dynamic specification for the animated visualizations using familiar map-reduce paradigm. In the context of deep learning training, this allows users to create on-demand custom visualizations such as per-layer gradient statistics change over time, display of sample predictions sorted by loss value and so on.


\section{Stream Generation Using Map-Reduce}
\subsection{Background}
There are many variants of the map-reduce model\cite{Afrati:2011:MER:1951365.1951367} and differences in various implementations. We will focus on the variant that is popular among data scientists and readily available in widely used programming languages such as Python.

The map-reduce paradigm consists of two higher order operators: \emph{map} and \emph{reduce}. The map operator accepts a function $M$ and a list of values $V$. The $M$ is applied to each value in $V$ to transform it to some other value or choose not to output any value, i.e.\ , the filter operation.

The reduce operator accepts a function $R$ and a list of values $V$. The $R$ processes each value in $V$ to produce an aggregated output value. For instance the operation of sum over a sequence can be done as reduce operation with $R$ that initializes aggregated value to $0$ and then consumes each value in the sequence to produce new aggregated value.



\subsection{Extending Map-Reduce}\label{map-reduce-ext}
While the map operator consumes a stream and outputs a stream, the reduce operator consumes stream and outputs an aggregated value instead of a stream. The reduce operator's output is not generated until the entire stream ends. In several of our scenarios, we rather desire that the reduce operator works on a group of contiguous values in the stream, aggregating values in that group and outputting a stream. For instance, we may want to compute the average duration for batches within each epoch and generate a stream with these averages as epochs progresses.

To achieve this, we introduce an extension to allow us leveraging the existing infrastructure and avoid need for entirely new domain specific language. In our extension, we simply require that each value in the stream is accompanied by an \emph{optional} binary value $B$ which when $true$ triggers the output from the reduce operator.

There are two advantages offered by this design:

\begin{enumerate}
  \item $B$ can be set at any time by the host process $P$ enabling many of our core scenarios trivially.
  \item $B$ can also be set by a client at any time. This enables the scenarios where the user dynamically defines the aggregation window. For example, the user may wish to view a metric averaged over every 5 minutes.
\end{enumerate}

\section{Implementation}
We implement our design using Python and other frameworks described in this section. We will be releasing our implementation as an open source cross-platform offering.

For networking stack we utilize the ZeroMQ library to implement publisher-subscriber model between the agent and the client. Out of the box, we offer implementations for MatplotLib as well as Plotly frameworks for various visualizations including line charts, histograms and image matrix. MatplotLib allows a variety of UX backends, many of which can run as native application or in Notebook interface for exploratory tasks. The Jupyter Lab allows transforming Notebook in to the user defined dashboards.

One of the key requirements in our system model is the implementation of the map-reduce extension described in Section \ref{map-reduce-ext}. We achieve this by implementing a component we call \emph{postable iterator}. The postable iterator allows to post input sequence of tuple $\{value, B\}$, where $B$ is group completion flag described in the Section \ref{map-reduce-ext}. The postable iterator then evaluates the map-reduce expression and returns the output value of map or reduce operator or signals the caller that no output was produced for the posted value.

One of the key difficulties in implementation using languages such as Python and frameworks such as ZeroMQ, MatplotLib, and Jupyter Notebook is managing the limitations imposed for multi-threading. 
We adopt the cooperative concurrency model with callbacks combining with the producer-consumer pattern to work around many of these limitations.

\section{Conclusion}
We described the design of a system that brings data streaming and map-reduce style queries to the domain of machine learning training for enabling the new scenarios of diagnosis and exploratory inspection. We identified several advantages of our system over currently popular systems, including the ability to perform interactive queries, dynamic construction of data flow pipelines, and decoupled adaptive visualizations as nodes in such pipelines. We plan to release our system as an open source cross-platform offering to help researchers and engineers perform diagnosis and exploratory tasks more efficiently for the deep learning training processes.

%
\begin{acks}
We would like to thank Susan Dumais for her guidance and advice on this project.
\end{acks}

%
\bibliographystyle{ACM-Reference-Format}
\bibliography{sample-base}


\begin{thebibliography}{26}


\ifx \showCODEN    \undefined \def \showCODEN     #1{\unskip}     \fi
\ifx \showDOI      \undefined \def \showDOI       #1{#1}\fi
\ifx \showISBNx    \undefined \def \showISBNx     #1{\unskip}     \fi
\ifx \showISBNxiii \undefined \def \showISBNxiii  #1{\unskip}     \fi
\ifx \showISSN     \undefined \def \showISSN      #1{\unskip}     \fi
\ifx \showLCCN     \undefined \def \showLCCN      #1{\unskip}     \fi
\ifx \shownote     \undefined \def \shownote      #1{#1}          \fi
\ifx \showarticletitle \undefined \def \showarticletitle #1{#1}   \fi
\ifx \showURL      \undefined \def \showURL       {\relax}        \fi
\providecommand\bibfield[2]{#2}
\providecommand\bibinfo[2]{#2}
\providecommand\natexlab[1]{#1}
\providecommand\showeprint[2][]{arXiv:#2}

\bibitem[\protect\citeauthoryear{??}{Vis}{[n. d.]}]%
        {VisualDL}
 \bibinfo{year}{[n. d.]}\natexlab{}.
\newblock \bibinfo{title}{Visualize your deep learning training and data
  flawlessly}.
\newblock
\newblock
\urldef\tempurl%
\url{https://github.com/PaddlePaddle/VisualDL}
\showURL{%
\tempurl}


\bibitem[\protect\citeauthoryear{Afrati, Borkar, Carey, Polyzotis, and
  Ullman}{Afrati et~al\mbox{.}}{2011}]%
        {Afrati:2011:MER:1951365.1951367}
\bibfield{author}{\bibinfo{person}{Foto~N. Afrati}, \bibinfo{person}{Vinayak
  Borkar}, \bibinfo{person}{Michael Carey}, \bibinfo{person}{Neoklis
  Polyzotis}, {and} \bibinfo{person}{Jeffrey~D. Ullman}.}
  \bibinfo{year}{2011}\natexlab{}.
\newblock \showarticletitle{Map-reduce Extensions and Recursive Queries}. In
  \bibinfo{booktitle}{\emph{Proceedings of the 14th International Conference on
  Extending Database Technology}} \emph{(\bibinfo{series}{EDBT/ICDT '11})}.
  \bibinfo{publisher}{ACM}, \bibinfo{address}{New York, NY, USA},
  \bibinfo{pages}{1--8}.
\newblock
\showISBNx{978-1-4503-0528-0}
\urldef\tempurl%
\url{https://doi.org/10.1145/1951365.1951367}
\showDOI{\tempurl}


\bibitem[\protect\citeauthoryear{Babu and Widom}{Babu and Widom}{2001}]%
        {Babu:2001:CQO:603867.603884}
\bibfield{author}{\bibinfo{person}{Shivnath Babu} {and}
  \bibinfo{person}{Jennifer Widom}.} \bibinfo{year}{2001}\natexlab{}.
\newblock \showarticletitle{Continuous Queries over Data Streams}.
\newblock \bibinfo{journal}{\emph{SIGMOD Rec.}} \bibinfo{volume}{30},
  \bibinfo{number}{3} (\bibinfo{date}{Sept.} \bibinfo{year}{2001}),
  \bibinfo{pages}{109--120}.
\newblock
\showISSN{0163-5808}
\urldef\tempurl%
\url{https://doi.org/10.1145/603867.603884}
\showDOI{\tempurl}


\bibitem[\protect\citeauthoryear{Bergstra, Bardenet, Bengio, and
  K\'{e}gl}{Bergstra et~al\mbox{.}}{2011}]%
        {NIPS2011_4443}
\bibfield{author}{\bibinfo{person}{James~S. Bergstra},
  \bibinfo{person}{R\'{e}mi Bardenet}, \bibinfo{person}{Yoshua Bengio}, {and}
  \bibinfo{person}{Bal\'{a}zs K\'{e}gl}.} \bibinfo{year}{2011}\natexlab{}.
\newblock \showarticletitle{Algorithms for Hyper-Parameter Optimization}.
\newblock In \bibinfo{booktitle}{\emph{Advances in Neural Information
  Processing Systems 24}}, \bibfield{editor}{\bibinfo{person}{J.~Shawe-Taylor},
  \bibinfo{person}{R.~S. Zemel}, \bibinfo{person}{P.~L. Bartlett},
  \bibinfo{person}{F.~Pereira}, {and} \bibinfo{person}{K.~Q. Weinberger}}
  (Eds.). \bibinfo{publisher}{Curran Associates, Inc.},
  \bibinfo{pages}{2546--2554}.
\newblock
\urldef\tempurl%
\url{http://papers.nips.cc/paper/4443-algorithms-for-hyper-parameter-optimization.pdf}
\showURL{%
\tempurl}


\bibitem[\protect\citeauthoryear{Catanzaro, Sundaram, and Keutzer}{Catanzaro
  et~al\mbox{.}}{2008}]%
        {Catanzaro2008AMR}
\bibfield{author}{\bibinfo{person}{Bryan Catanzaro}, \bibinfo{person}{Narayanan
  Sundaram}, {and} \bibinfo{person}{Kurt Keutzer}.}
  \bibinfo{year}{2008}\natexlab{}.
\newblock \showarticletitle{A map reduce framework for programming graphics
  processors}.
\newblock


\bibitem[\protect\citeauthoryear{Choo and Liu}{Choo and Liu}{2018}]%
        {Choo2018}
\bibfield{author}{\bibinfo{person}{Jaegul Choo} {and} \bibinfo{person}{Shixia
  Liu}.} \bibinfo{year}{2018}\natexlab{}.
\newblock \showarticletitle{Visual Analytics for Explainable Deep Learning}.
\newblock \bibinfo{journal}{\emph{{IEEE} Computer Graphics and Applications}}
  \bibinfo{volume}{38}, \bibinfo{number}{4} (\bibinfo{date}{jul}
  \bibinfo{year}{2018}), \bibinfo{pages}{84--92}.
\newblock
\urldef\tempurl%
\url{https://doi.org/10.1109/mcg.2018.042731661}
\showDOI{\tempurl}


\bibitem[\protect\citeauthoryear{Dean and Ghemawat}{Dean and Ghemawat}{2008}]%
        {Dean2008}
\bibfield{author}{\bibinfo{person}{Jeffrey Dean} {and} \bibinfo{person}{Sanjay
  Ghemawat}.} \bibinfo{year}{2008}\natexlab{}.
\newblock \showarticletitle{{MapReduce}}.
\newblock \bibinfo{journal}{\emph{Commun. ACM}} \bibinfo{volume}{51},
  \bibinfo{number}{1} (\bibinfo{date}{jan} \bibinfo{year}{2008}),
  \bibinfo{pages}{107}.
\newblock
\urldef\tempurl%
\url{https://doi.org/10.1145/1327452.1327492}
\showDOI{\tempurl}


\bibitem[\protect\citeauthoryear{(Ed.), Bajaj, and Wiley}{(Ed.)
  et~al\mbox{.}}{1998}]%
        {(Ed.)98datavisualization}
\bibfield{author}{\bibinfo{person}{Chandrajit~Bajaj (Ed.)},
  \bibinfo{person}{Chandrajit Bajaj}, {and} \bibinfo{person}{C~Fl~John Wiley}.}
  \bibinfo{year}{1998}\natexlab{}.
\newblock \bibinfo{title}{Data Visualization Techniques}.
\newblock
\newblock


\bibitem[\protect\citeauthoryear{Ekanayake, Pallickara, and Fox}{Ekanayake
  et~al\mbox{.}}{2008}]%
        {Ekanayake2008}
\bibfield{author}{\bibinfo{person}{Jaliya Ekanayake}, \bibinfo{person}{Shrideep
  Pallickara}, {and} \bibinfo{person}{Geoffrey Fox}.}
  \bibinfo{year}{2008}\natexlab{}.
\newblock \showarticletitle{{MapReduce} for Data Intensive Scientific
  Analyses}. In \bibinfo{booktitle}{\emph{2008 {IEEE} Fourth International
  Conference on {eScience}}}. \bibinfo{publisher}{{IEEE}}.
\newblock
\urldef\tempurl%
\url{https://doi.org/10.1109/escience.2008.59}
\showDOI{\tempurl}


\bibitem[\protect\citeauthoryear{Ellis}{Ellis}{2014}]%
        {ellis2014real}
\bibfield{author}{\bibinfo{person}{B. Ellis}.} \bibinfo{year}{2014}\natexlab{}.
\newblock \bibinfo{booktitle}{\emph{Real-Time Analytics: Techniques to Analyze
  and Visualize Streaming Data}}.
\newblock \bibinfo{publisher}{Wiley}.
\newblock
\showISBNx{9781118838020}
\urldef\tempurl%
\url{https://books.google.com/books?id=DFnOAwAAQBAJ}
\showURL{%
\tempurl}


\bibitem[\protect\citeauthoryear{Few}{Few}{2006}]%
        {Few:2006:IDD:1206491}
\bibfield{author}{\bibinfo{person}{Stephen Few}.}
  \bibinfo{year}{2006}\natexlab{}.
\newblock \bibinfo{booktitle}{\emph{Information Dashboard Design: The Effective
  Visual Communication of Data}}.
\newblock \bibinfo{publisher}{O'Reilly Media, Inc.}
\newblock
\showISBNx{0596100167}


\bibitem[\protect\citeauthoryear{Gates, Natkovich, Chopra, Kamath,
  Narayanamurthy, Olston, Reed, Srinivasan, and Srivastava}{Gates
  et~al\mbox{.}}{2009}]%
        {Gates:2009:BHD:1687553.1687568}
\bibfield{author}{\bibinfo{person}{Alan~F. Gates}, \bibinfo{person}{Olga
  Natkovich}, \bibinfo{person}{Shubham Chopra}, \bibinfo{person}{Pradeep
  Kamath}, \bibinfo{person}{Shravan~M. Narayanamurthy},
  \bibinfo{person}{Christopher Olston}, \bibinfo{person}{Benjamin Reed},
  \bibinfo{person}{Santhosh Srinivasan}, {and} \bibinfo{person}{Utkarsh
  Srivastava}.} \bibinfo{year}{2009}\natexlab{}.
\newblock \showarticletitle{Building a High-level Dataflow System on Top of
  Map-Reduce: The Pig Experience}.
\newblock \bibinfo{journal}{\emph{Proc. VLDB Endow.}} \bibinfo{volume}{2},
  \bibinfo{number}{2} (\bibinfo{date}{Aug.} \bibinfo{year}{2009}),
  \bibinfo{pages}{1414--1425}.
\newblock
\showISSN{2150-8097}
\urldef\tempurl%
\url{https://doi.org/10.14778/1687553.1687568}
\showDOI{\tempurl}


\bibitem[\protect\citeauthoryear{Hastie and Tibshirani}{Hastie and
  Tibshirani}{1986}]%
        {Hastie1986}
\bibfield{author}{\bibinfo{person}{Trevor Hastie} {and} \bibinfo{person}{Robert
  Tibshirani}.} \bibinfo{year}{1986}\natexlab{}.
\newblock \showarticletitle{Generalized Additive Models}.
\newblock \bibinfo{journal}{\emph{Statist. Sci.}} \bibinfo{volume}{1},
  \bibinfo{number}{3} (\bibinfo{date}{aug} \bibinfo{year}{1986}),
  \bibinfo{pages}{297--310}.
\newblock
\urldef\tempurl%
\url{https://doi.org/10.1214/ss/1177013604}
\showDOI{\tempurl}


\bibitem[\protect\citeauthoryear{Liu, Wang, Liu, and Zhu}{Liu
  et~al\mbox{.}}{2017}]%
        {Liu2017}
\bibfield{author}{\bibinfo{person}{Shixia Liu}, \bibinfo{person}{Xiting Wang},
  \bibinfo{person}{Mengchen Liu}, {and} \bibinfo{person}{Jun Zhu}.}
  \bibinfo{year}{2017}\natexlab{}.
\newblock \showarticletitle{Towards better analysis of machine learning models:
  A visual analytics perspective}.
\newblock \bibinfo{journal}{\emph{Visual Informatics}} \bibinfo{volume}{1},
  \bibinfo{number}{1} (\bibinfo{date}{mar} \bibinfo{year}{2017}),
  \bibinfo{pages}{48--56}.
\newblock
\urldef\tempurl%
\url{https://doi.org/10.1016/j.visinf.2017.01.006}
\showDOI{\tempurl}


\bibitem[\protect\citeauthoryear{Logre and D{\'e}ry-Pinna}{Logre and
  D{\'e}ry-Pinna}{2018}]%
        {Logre:2018:MSV:3233739.3229096}
\bibfield{author}{\bibinfo{person}{Ivan Logre} {and}
  \bibinfo{person}{Anne-Marie D{\'e}ry-Pinna}.}
  \bibinfo{year}{2018}\natexlab{}.
\newblock \showarticletitle{MDE in Support of Visualization Systems Design: A
  Multi-Staged Approach Tailored for Multiple Roles}.
\newblock \bibinfo{journal}{\emph{Proc. ACM Hum.-Comput. Interact.}}
  \bibinfo{volume}{2}, \bibinfo{number}{EICS}, Article \bibinfo{articleno}{14}
  (\bibinfo{date}{June} \bibinfo{year}{2018}), \bibinfo{numpages}{17}~pages.
\newblock
\showISSN{2573-0142}
\urldef\tempurl%
\url{https://doi.org/10.1145/3229096}
\showDOI{\tempurl}


\bibitem[\protect\citeauthoryear{Mourlas}{Mourlas}{2009}]%
        {mourlas2009intelligent}
\bibfield{author}{\bibinfo{person}{Constantinos Mourlas}.}
  \bibinfo{year}{2009}\natexlab{}.
\newblock \bibinfo{booktitle}{\emph{Intelligent user interfaces : adaptation
  and personalization systems and technologies}}.
\newblock \bibinfo{publisher}{Information Science Reference},
  \bibinfo{address}{Hershey, PA}.
\newblock
\showISBNx{9781605660325}


\bibitem[\protect\citeauthoryear{Nazemi}{Nazemi}{2016}]%
        {Nazemi2016}
\bibfield{author}{\bibinfo{person}{Kawa Nazemi}.}
  \bibinfo{year}{2016}\natexlab{}.
\newblock \bibinfo{booktitle}{\emph{Adaptive Semantics Visualization}}.
\newblock \bibinfo{publisher}{Springer International Publishing}.
\newblock
\urldef\tempurl%
\url{https://doi.org/10.1007/978-3-319-30816-6}
\showDOI{\tempurl}


\bibitem[\protect\citeauthoryear{Pavlo, Paulson, Rasin, Abadi, DeWitt, Madden,
  and Stonebraker}{Pavlo et~al\mbox{.}}{2009}]%
        {Pavlo:2009:CAL:1559845.1559865}
\bibfield{author}{\bibinfo{person}{Andrew Pavlo}, \bibinfo{person}{Erik
  Paulson}, \bibinfo{person}{Alexander Rasin}, \bibinfo{person}{Daniel~J.
  Abadi}, \bibinfo{person}{David~J. DeWitt}, \bibinfo{person}{Samuel Madden},
  {and} \bibinfo{person}{Michael Stonebraker}.}
  \bibinfo{year}{2009}\natexlab{}.
\newblock \showarticletitle{A Comparison of Approaches to Large-scale Data
  Analysis}. In \bibinfo{booktitle}{\emph{Proceedings of the 2009 ACM SIGMOD
  International Conference on Management of Data}}
  \emph{(\bibinfo{series}{SIGMOD '09})}. \bibinfo{publisher}{ACM},
  \bibinfo{address}{New York, NY, USA}, \bibinfo{pages}{165--178}.
\newblock
\showISBNx{978-1-60558-551-2}
\urldef\tempurl%
\url{https://doi.org/10.1145/1559845.1559865}
\showDOI{\tempurl}


\bibitem[\protect\citeauthoryear{Plale and Schwan}{Plale and Schwan}{2003}]%
        {Plale2003}
\bibfield{author}{\bibinfo{person}{B. Plale} {and} \bibinfo{person}{K.
  Schwan}.} \bibinfo{year}{2003}\natexlab{}.
\newblock \showarticletitle{Dynamic querying of streaming data with the {dQUOB}
  system}.
\newblock \bibinfo{journal}{\emph{{IEEE} Transactions on Parallel and
  Distributed Systems}} \bibinfo{volume}{14}, \bibinfo{number}{4}
  (\bibinfo{date}{apr} \bibinfo{year}{2003}), \bibinfo{pages}{422--432}.
\newblock
\urldef\tempurl%
\url{https://doi.org/10.1109/tpds.2003.1195413}
\showDOI{\tempurl}


\bibitem[\protect\citeauthoryear{Roberts}{Roberts}{2007}]%
        {Roberts2007}
\bibfield{author}{\bibinfo{person}{Jonathan~C. Roberts}.}
  \bibinfo{year}{2007}\natexlab{}.
\newblock \showarticletitle{State of the Art: Coordinated \& Multiple Views in
  Exploratory Visualization}. In \bibinfo{booktitle}{\emph{Fifth International
  Conference on Coordinated and Multiple Views in Exploratory Visualization
  ({CMV} 2007)}}. \bibinfo{publisher}{{IEEE}}.
\newblock
\urldef\tempurl%
\url{https://doi.org/10.1109/cmv.2007.20}
\showDOI{\tempurl}


\bibitem[\protect\citeauthoryear{Steele}{Steele}{1995}]%
        {Steele1995}
\bibfield{author}{\bibinfo{person}{Guy~L. Steele}.}
  \bibinfo{year}{1995}\natexlab{}.
\newblock \showarticletitle{Parallelism in Lisp}.
\newblock \bibinfo{journal}{\emph{{ACM} {SIGPLAN} Lisp Pointers}}
  \bibinfo{volume}{{VIII}}, \bibinfo{number}{2} (\bibinfo{date}{may}
  \bibinfo{year}{1995}), \bibinfo{pages}{1--14}.
\newblock
\urldef\tempurl%
\url{https://doi.org/10.1145/224133.224134}
\showDOI{\tempurl}


\bibitem[\protect\citeauthoryear{Stockinger, Shalf, Wu, and Bethel}{Stockinger
  et~al\mbox{.}}{[n. d.]}]%
        {Stockinger}
\bibfield{author}{\bibinfo{person}{K. Stockinger}, \bibinfo{person}{J. Shalf},
  \bibinfo{person}{Kesheng Wu}, {and} \bibinfo{person}{E.W. Bethel}.}
  \bibinfo{year}{[n. d.]}\natexlab{}.
\newblock \showarticletitle{Query-Driven Visualization of Large Data Sets}. In
  \bibinfo{booktitle}{\emph{{VIS} 05. {IEEE} Visualization, 2005.}}
  \bibinfo{publisher}{{IEEE}}.
\newblock
\urldef\tempurl%
\url{https://doi.org/10.1109/visual.2005.1532792}
\showDOI{\tempurl}


\bibitem[\protect\citeauthoryear{Sung, Kim, Jo, Yang, Kim, Lausen, Kim, Lee,
  Kwak, Ha, and Kim}{Sung et~al\mbox{.}}{2017}]%
        {DBLP:journals/corr/abs-1712-05902}
\bibfield{author}{\bibinfo{person}{Nako Sung}, \bibinfo{person}{Minkyu Kim},
  \bibinfo{person}{Hyunwoo Jo}, \bibinfo{person}{Youngil Yang},
  \bibinfo{person}{Jingwoong Kim}, \bibinfo{person}{Leonard Lausen},
  \bibinfo{person}{Youngkwan Kim}, \bibinfo{person}{Gayoung Lee},
  \bibinfo{person}{Dong{-}Hyun Kwak}, \bibinfo{person}{Jung{-}Woo Ha}, {and}
  \bibinfo{person}{Sunghun Kim}.} \bibinfo{year}{2017}\natexlab{}.
\newblock \showarticletitle{{NSML:} {A} Machine Learning Platform That Enables
  You to Focus on Your Models}.
\newblock \bibinfo{journal}{\emph{CoRR}}  \bibinfo{volume}{abs/1712.05902}
  (\bibinfo{year}{2017}).
\newblock
\showeprint[arxiv]{1712.05902}
\urldef\tempurl%
\url{http://arxiv.org/abs/1712.05902}
\showURL{%
\tempurl}


\bibitem[\protect\citeauthoryear{Traub, Steenbergen, Grulich, Rabl, and
  Markl}{Traub et~al\mbox{.}}{2017}]%
        {Traub2017}
\bibfield{author}{\bibinfo{person}{Jonas Traub}, \bibinfo{person}{Nikolaas
  Steenbergen}, \bibinfo{person}{Philipp Grulich}, \bibinfo{person}{Tilmann
  Rabl}, {and} \bibinfo{person}{Volker Markl}.}
  \bibinfo{year}{2017}\natexlab{}.
\newblock \bibinfo{title}{I²: Interactive Real-Time Visualization for
  Streaming Data}.
\newblock
\newblock
\urldef\tempurl%
\url{https://doi.org/10.5441/002/edbt.2017.61}
\showDOI{\tempurl}


\bibitem[\protect\citeauthoryear{Wongsuphasawat, Smilkov, Wexler, Wilson, Mane,
  Fritz, Krishnan, Viegas, and Wattenberg}{Wongsuphasawat
  et~al\mbox{.}}{2018}]%
        {Wongsuphasawat2018}
\bibfield{author}{\bibinfo{person}{Kanit Wongsuphasawat},
  \bibinfo{person}{Daniel Smilkov}, \bibinfo{person}{James Wexler},
  \bibinfo{person}{Jimbo Wilson}, \bibinfo{person}{Dandelion Mane},
  \bibinfo{person}{Doug Fritz}, \bibinfo{person}{Dilip Krishnan},
  \bibinfo{person}{Fernanda~B. Viegas}, {and} \bibinfo{person}{Martin
  Wattenberg}.} \bibinfo{year}{2018}\natexlab{}.
\newblock \showarticletitle{Visualizing Dataflow Graphs of Deep Learning Models
  in {TensorFlow}}.
\newblock \bibinfo{journal}{\emph{{IEEE} Transactions on Visualization and
  Computer Graphics}} \bibinfo{volume}{24}, \bibinfo{number}{1}
  (\bibinfo{date}{jan} \bibinfo{year}{2018}), \bibinfo{pages}{1--12}.
\newblock
\urldef\tempurl%
\url{https://doi.org/10.1109/tvcg.2017.2744878}
\showDOI{\tempurl}


\bibitem[\protect\citeauthoryear{You, Zhang, Hsieh, Demmel, and Keutzer}{You
  et~al\mbox{.}}{2018}]%
        {You:2018:ITM:3225058.3225069}
\bibfield{author}{\bibinfo{person}{Yang You}, \bibinfo{person}{Zhao Zhang},
  \bibinfo{person}{Cho-Jui Hsieh}, \bibinfo{person}{James Demmel}, {and}
  \bibinfo{person}{Kurt Keutzer}.} \bibinfo{year}{2018}\natexlab{}.
\newblock \showarticletitle{ImageNet Training in Minutes}. In
  \bibinfo{booktitle}{\emph{Proceedings of the 47th International Conference on
  Parallel Processing}} \emph{(\bibinfo{series}{ICPP 2018})}.
  \bibinfo{publisher}{ACM}, \bibinfo{address}{New York, NY, USA}, Article
  \bibinfo{articleno}{1}, \bibinfo{numpages}{10}~pages.
\newblock
\showISBNx{978-1-4503-6510-9}
\urldef\tempurl%
\url{https://doi.org/10.1145/3225058.3225069}
\showDOI{\tempurl}


\end{thebibliography}

%

\end{document}